# Helpful Neighbors:
# Leveraging Neighbors in Geographic Feature Pronunciation

**Llion Jones**[†]    **Richard Sproat**[†]    **Haruko Ishikawa**[†]    **Alexander Gutkin**[‡]
[†]Google Japan    [‡]Google UK
{llion,rws,ishikawa,agutkin}@google.com

## Abstract

If one sees the place name *Houston Mercer Dog Run* in New York, how does one know how to pronounce it? Assuming one knows that *Houston* in New York is pronounced /ˈhaʊstən/ and not like the Texas city (/ˈhjuːstən/), then one can probably guess that /ˈhaʊstən/ is also used in the name of the dog park. We present a novel architecture that learns to use the pronunciations of neighboring names in order to guess the pronunciation of a given target feature. Applied to Japanese place names, we demonstrate the utility of the model to finding and proposing corrections for errors in Google Maps.

To demonstrate the utility of this approach to structurally similar problems, we also report on an application to a totally different task: *Cognate reflex prediction* in comparative historical linguistics. A version of the code has been open-sourced.[1]

## 1 Introduction

In many parts of the world, pronunciation of toponyms and establishments can require local knowledge. Many visitors to New York, for example, get tripped up by *Houston Street*, which they assume is pronounced the same as the city in Texas. If they do not know how to pronounce *Houston Street*, they would likely also not know how to pronounce the nearby *Houston Mercer Dog Run*. But if one knows one, that can (usually) be used as a clue to how to pronounce the other.

Before we proceed further, a bit of terminology. Technically, the term *toponym* refers to the name of a geographical or administrative feature, such as a river, lake, town or state. In most of what follows, we will use the term *feature* to refer to these and other entities such as roads, buildings, schools etc. In practice we will not make a major distinction between the two, but since there is a sense in which toponyms are more basic, and the names of the more general features are often derived from a toponym (as in the *Houston Mercer Dog Run* example above), we will retain the distinction where it is needed.

While features cause not infrequent problems in the US, they become a truly serious issue in Japan. Japan is notorious for having toponyms whose pronunciation is so unexpected that even native speakers may not know how to pronounce a given case. Most toponyms in Japanese are written in kanji (Chinese characters) with a possible intermixing of one of the two syllabaries, hiragana or katakana. Thus 上野 *Ueno* is entirely in kanji; 虎ノ門 *Tora no mon* has two kanji and one katakana symbol (the second); and 吹割の滝 *Fukiwari Waterfalls* has three kanji and one hiragana symbol (the third). Features more generally tend to have more characters in one of the syllabaries—especially katakana if, for example, the feature is a building that includes the name of a company as part of its name.

The syllabaries are basically phonemic scripts so there is generally no ambiguity in how to pronounce those portions of names, but kanji present a serious problem in that the pronunciation of a kanji string in a toponym is frequently something one just has to know. To take the example 上野 *Ueno* above, that pronunciation (for the well-known area in Tokyo) is indeed the most common one, but there are places in Japan with the same spelling but with pronunciations such as *Uwano*, *Kamino*, *Wano*, among others.[2] It is well-known that many kanji have both a *native* (*kun*) Japanese pronunciation (e.g. 山 *yama* 'mountain') as well as one or more Chinese-derived *on* pronunciations (e.g. 山

---

[1]https://github.com/google-research/google-research/tree/master/cognate_inpaint_neighbors

[2]Different pronunciations of kanji are often referred to as *readings*, but in this paper we will use the more general term *pronunciation*.

*san* 'mountain'), but the issue with toponyms goes well beyond this since there are *nanori* pronunciations of kanji that are only found in names (Ogihara, 2021): 山 also has the *nanori* pronunciation *taka*, for example. The *kun-on-nanori* variants relate to an important property of how kanji are used in Japanese: among all modern writing systems, the Japanese use of kanji comes closest to being *semasiographic*—i.e. representing meaning rather than specific morphemes. The common toponym component *kawa* 'river', is usually written 川, but can also be written as 河, which also means 'river'. That kanji in turn has other pronunciations, such as *kō*, a Sino-Japanese word for 'river'. This freedom to spell words with a range of kanji that have the same meaning, or to read kanji with any of a number of morphemes having the same meaning, is a particular characteristic of Japanese. Thus, while reading place names can be tricky in many parts of the world, the problem is particularly acute in Japan.

Since the variation is largely unpredictable, one therefore simply needs to know for a given toponym what the pronunciation is. But once one knows, for instance, that a name written 上野 is read as *Uwano*, as with the *Houston* case, one ought to be able to deduce that in the name of the local 上野第1公園 'Uwano First Public Park', this is read as *Uwano* and not *Ueno*. If one's digital assistant is reading this name to you, or needs to understand your pronunciation of the name, it needs to know the correct pronunciation. While one might expect a complete and correct maps database to have all of this information correctly entered, in practice maps data contain many errors, especially for less frequently accessed features.

In this paper we propose a model that learns to use information from the geographical context to guide the pronunciation of features. We demonstrate its application to detecting and correcting errors in Google Maps. In addition, in Section 8 we show that the model can be applied to a different but structurally similar problem, namely the problem of *cognate reflex prediction* in comparative historical linguistics. In this case the 'neighbors' are related word forms in a set of languages from a given language family, and the pronunciation to be predicted is the corresponding form in a language from the same family.

## 2 Background

Pronouncing written geographical feature names involves a combination of text normalization (if the names contain expressions such as numbers or abbreviations), and word pronunciation, often termed "grapheme-to-phoneme conversion". Both of these are typically cast as sequence-to-sequence problems, and neural approaches to both are now common. For neural approaches to grapheme-to-phoneme conversion see (Yao and Zweig, 2015; Rao et al., 2015; Toshniwal and Livescu, 2016; Peters et al., 2017; Yolchuyeva et al., 2019), and for text normalization see (Sproat and Jaitly, 2017; Zhang et al., 2019; Yolchuyeva et al., 2018; Pramanik and Hussain, 2019; Mansfield et al., 2019; Kawamura et al., 2020; Tran and Bui, 2021). For languages that use the Chinese script, grapheme-to-phoneme conversion may benefit from the fact that Chinese characters can mostly be decomposed into a component that relates to the meaning of the character and another that relates to the pronunciation. The latter information is potentially useful, in particular in Chinese and in the Sino-Japanese readings of characters in Japanese. Recent neural models that have taken advantage of this include (Dai and Cai, 2017; Nguyen et al., 2020). On the other hand, it should be pointed out that other more 'brute force' decompositions of characters seem to be useful. Thus Yu et al. (2020) propose a *byte decomposition* for (UTF-8) character encodings for a model that covers a wide variety of languages, including Chinese and Japanese.

The above approaches generally treat the problem in isolation in the sense that the problem is cast as one where the task is to predict a pronunciation independent of context. Different pronunciations for the same string in different linguistic contexts comes under the rubric of *homograph disambiguation*, and there is a long tradition of work in this area; for an early example see (Yarowsky, 1996) and for a recent incarnation see (Gorman et al., 2018). Not surprisingly, there has been recent interest in neural models for predicting homograph pronunciations: see (Park and Lee, 2020; Shi et al., 2021) for recent examples focused on Mandarin.

The present task is different, since what disambiguates the possible pronunciations of Japanese features is not generally linguistic, but geographical context, which can be thought of as a way of *biasing* the decision as to which pronunciation to use, given evidence from the local context. Our approach is similar in spirit to that of Pundak et al. (2018), who propose the use of a *bias-encoder* in a "listen-attend-and-spell" (Chan et al., 2016) Automatic Speech Recognition architecture. The bias encoder takes a set of "bias phrases", which can be used to guide the model towards a particular decoding. Pundak et al. (2018)'s model is shown schematically in Figure 1.

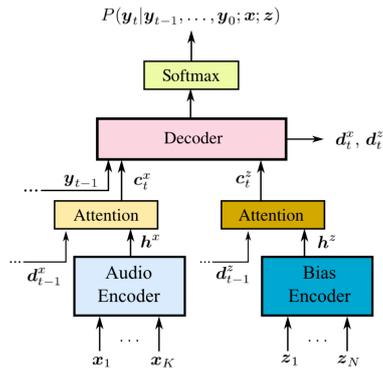

Figure 1: The biasing LAS model from (Pundak et al., 2018), Figure 1a.

## 3 Data

Features in Google Maps are stored in a data representation that includes a variety of information about each feature including: its location as a bounding box in latitude-longitude; the type of the feature—street, building, municipality, topographic feature, etc.; name(s) of the feature in the native language as well as in many (mostly automatically generated) transliterations; an address if there is an address associated with this feature; road signs that may be associated; and so forth. Each feature is identified with a unique hexadecimal feature id. Features may have additional names besides the primary names. For example in English, street names are often abbreviated (*Main St.*) and these abbreviations are typically expanded (*Main Street*) as an additional name. Many Japanese features have pronunciations of the names added as additional names in katakana. Some of these have been carefully hand curated, but many were generated automatically and are therefore potentially errorful, as we will see. Since the katakana version is used as the basis for transliterations into other languages, localized pronunciations for text-to-speech, as well as search suggestions, it is important that it be correct.

We started by extracting from the database all features that include a broad (but not exhaustive) set of feature types from a bounding box that covers the four main islands of Japan. We then extracted feature summaries for names that included both kanji original names, and katakana renditions. These summaries include the feature name, the hiragana version of the name converted from katakana, and the bounding box for the feature. We then find, for each feature in the feature summaries, a *bucket* of other features that are within a given radius (10 kilometers in our experiments). Then, for each feature in each bucket, we designate that feature a *target* feature, and we build *neighborhoods* around that feature. We attempt for each feature, to find *interesting* neighboring features whose name shares a kanji bigram with the target feature's name. The intuition here is that a feature that is likely to be useful in determining the pronunciation of another feature should be nearby geographically, and should share at least some of the name. In any case we cap the number of 'non-interesting' neighbors to a limit—5 in our experiments. This means that some neighborhoods will have target features that lack useful neighbors; this is a realistic situation in that while it is *often* the case that one can find hints for a name's pronunciation in the immediate neighbors, it is not *always* the case. While such neighborhoods are not useful from the point of view of neighbor-based evidence for a target feature's pronunciation, they still provide useful data for training the target sequence-to-sequence model. Our final dataset consists of about 2.7M feature neighborhoods, including the information from the summary for each target feature as described above, the associated neighboring features and their summaries, along with the distance (in kilometers) from the target feature. Figure 2 shows parts of one such neighborhood.

## 4 Model

Despite the differences noted above, the problem we are interested in can still be characterized at its core as a sequence-to-sequence problem. The input is a sequence of tokens representing the feature name in its original Japanese written form. The output is a sequence of hiragana characters representing the correct pronunciation. The difference between this and a more conventional sequence-to-sequence problem is that we provide additional biasing information in the form of geographical neighbors, such as their pronunciation and geo-graphical location. This neighbor information is provided as additional input sequences to aid the model in making its prediction. In our experiments, we limit the number of neighbors to at most 30 (it is usually much less than this), each consisting of two sequences, namely the neighbor's name and the corresponding pronunciation.

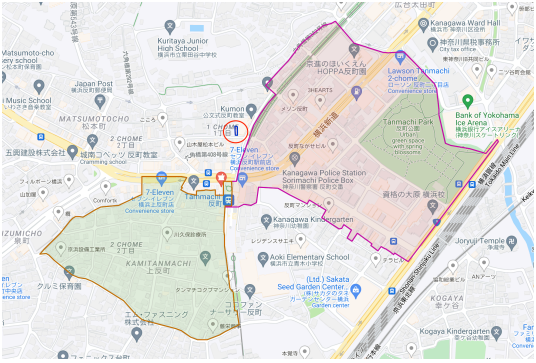

| Main | Name | セラヴィ反町 |
| | Pron | *seravi sorimachi* |
| | | (i.e. *C'est la Vie ...*) |
| Neigh: | Name | 反町 |
| | | *tanmachi* |
| | | (pink area on map) |
| Neigh: | Name | 上反町 |
| | | *kamitanmachi* |
| | | (green area on map) |

Figure 2: A small example of a neighborhood. The store, circled on the map, has a pronunciation listed as *C'est la Vie Sorimachi*, but the neighboring areas are *Tanmachi* and *Kamitanmachi*. *Sorimachi* is therefore wrong.

### 4.1 Model architecture

Due to many recent successes in other NLP applications, we experiment with a transformer model (Vaswani et al., 2017). Our transformer model uses a standard encoder-decoder architecture as the backbone. The inputs to the model are the input name with unknown pronunciation $x_{\text{inp}}$, the neighbor names $x_{\text{name}}$ (of length name_len) and associated pronunciations $x_{\text{pron}}$ (of length pron_len). First, these input tokens are embedded with size emb_size. The embeddings are then shared between the feature names and the pronunciations. i.e. the same embeddings are used for the input name tokens and the neighbor tokens, and similarly between the target pronunciation (decoder output) and the neighbors' pronunciations:

$$\text{emb}_{\text{inp}} = \text{Embed}_{\text{name}}(x_{\text{inp}}),$$
$$\text{emb}_{\text{name}} = \text{Embed}_{\text{name}}(x_{\text{name}}),$$
$$\text{emb}_{\text{pron}} = \text{Embed}_{\text{pron}}(x_{\text{pron}}).$$

These embedded tokens are then processed separately by the neighbor encoder. No parameters are shared between these encoders, or with the decoder:

$$h_{\text{inp}} = \text{Encoder}_{\text{inp}}(\text{emb}_{\text{inp}}),$$
$$h_{\text{name}} = \text{Encoder}_{\text{name}}(\text{emb}_{\text{name}}),$$
$$h_{\text{pron}} = \text{Encoder}_{\text{pron}}(\text{emb}_{\text{pron}}).$$

Since each example has nneigh neighbors, $h_{\text{inp}}$ is of shape [inp_size, emb_size] but the processed neighbor spelling and pronunciation inputs are of size [nneigh, name_len, emb_size] and [nneigh, pron_len, emb_size].

One of the simplest ways to incorporate the neighboring information is to concatenate the feature names and pronunciation embeddings into the main input sequence, allowing the transformer to attend directly to all the relevant information. Unfortunately, this is not possible with a vanilla transformer with a quadratic attention mechanism if we want to attend to, say, 30 neighbors. In our experiments name_len is set to 20 and pron_len is set to 40, yielding $(20 + 40) \times 30 = 1800$ input tokens, far too many for a vanilla transformer decoder to attend to. To mitigate against this we average the encoder outputs to give a single vector per neighbor to attend to:

$$s_{\text{name}} = \text{Ave}(h_{\text{name}}),$$
$$s_{\text{pron}} = \text{Ave}(h_{\text{pron}}),$$
$$c = \text{Concat}(h_{\text{inp}}, s_{\text{name}}, s_{\text{pron}}).$$

The vectors are concatenated along the neighbor dimension to give a sequence of size [inp_len+2*nneigh, emb_size]. Optionally, if embeddings representing the latitudinal and longitudinal position of the feature (which we refer to as Lat-Long embeddings, discussed later) are used then these are also concatenated here. This input sequence is then concatenated to the encoder output and is attended over by the transformer decoder. There are no positional embeddings added to this sequence, so they are unordered from the point of view of decoder attention. Therefore, we help the decoder match the neighbor names to their corresponding neighbor pronunciation by adding *source tokens* (Johnson et al., 2017) to the sequence. The same source token is added to matching names and pronunciation inputs. The specific hyperparameters used for all the transformer stacks are shown in Table 1.[3]

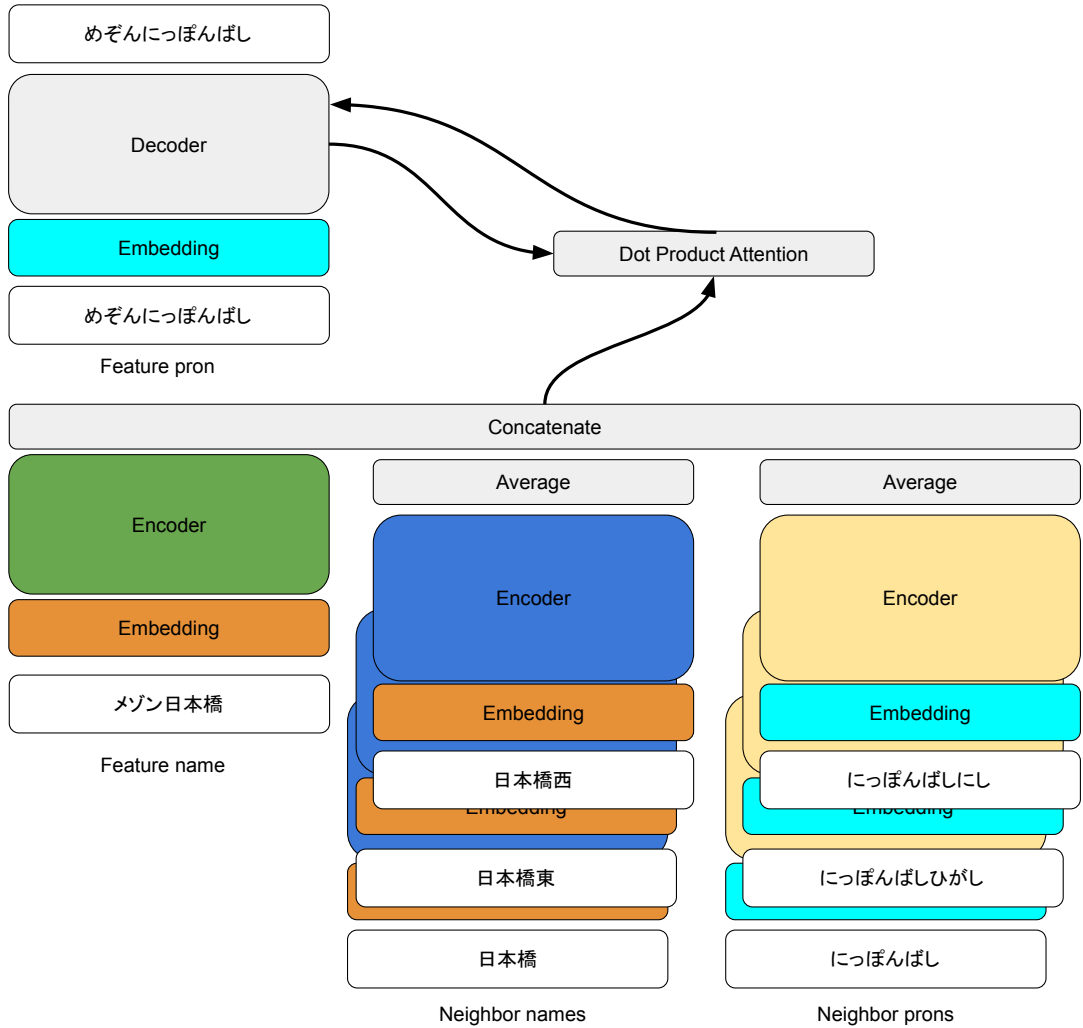

Figure 3: The transformer model, showing how the main feature and neighbor features are encoded. Colors for the embeddings and encoders reflect the shared parameters for the transformer model. Example shown is メゾン日本橋 *mezon nipponbashi*, and some neighboring features 日本橋 *nipponbashi*, 日本橋西 *nipponbashi nishi* and 日本橋東 *nipponbashi higashi*.

To combat overfitting, several types of dropout were employed. As in (Vaswani et al., 2017) we use input-dropout, where entire input embeddings can be dropped. We further use ReLU-dropout, dropping activations in the feed-forward layer after applying the ReLU non-linearity. Finally, we use attention-dropout, which is applied to the output of the attention layers. Additionally, dropout is applied to the auxiliary neighbor information, which means that a given neighbor's name or pronunciation, as well as the Lat-Long embedding, has a 10% chance of being dropped entirely in a train-ing example. The model can be configured to use neighbor information or not. We show below that the model benefits from neighbor information if it is available.

---
[3]Most feature names can be covered by 3,000 characters (Satō, 1985), so an input vocabulary of 4,710 kanji and other characters is a reasonable size for an industrial-scale maps database.

| Beam Search Size: | 8 |
|---|---|
| Number of layers: | 4 |
| # attention heads: | 8 |
| Token embedding size: | 256 |
| Hidden size: | 256 |
| Dropout: | 0.1 |
| Label smoothing: | 0.2 |
| Lat-Long grid size: | 100 |
| Input vocab size: | 4,710 |
| Output vocab size: | 427 |

Table 1: Hyperparameters for transformer stacks.

### 4.2 Lat-Long Embeddings

Some neighborhoods lack clues to pronunciation of the target feature. However, pronunciation of names is to some extent influenced by region, so the model might be able to deduce the pronunciation if given latitude/longitude coordinates of the main feature. We thus added embeddings to represent this information. An $n$ by $n$ grid was placed over Japan. Simply assigning a separate embedding to each square would require many embeddings and might slow the training. Also, due to Japan's shape, many embeddings would be in the sea and thus unused. Thus, rather than having $n^2$ embeddings, we treated each dimension separately resulting in $2n$ embeddings, each of size |emb_size/2|. The separate longitude and latitude embeddings for a given square are then concatenated together, and given to the decoder as an additional auxiliary input. Experiments showed that this configuration both trained faster and reduced overfitting.

### 4.3 Overfitting

One of the main challenges with training the model was overfitting. The reason for this was that it was known that there are incorrect pronunciations in the data and since we wanted to use the model to find errors, including ones in the training data, 100% accuracy on the training set was actually undesirable. A few techniques were used to combat overfitting. As well as the heavy use of dropout, label smoothing was set at 0.2, encouraging the model to be less confident about outliers. Since *source tokens* were added to the neighbor information, this made it easier for the model to memorize locations from their neighbor arrangements, so to mitigate against this the neighbors were shuffled within a batch before being processed by the model.[4] Also, care had to be taken to balance the size of the lat-long grid, between providing a useful clue to location, and allowing memorization of the location if the grid was too fine.

To assess potential overfitting during training, we created a small **golden** set of 2,008 high con-fidence pronunciations from the human evaluations that we ran while developing the model (Section 6). The distribution of these examples is very skewed with respect to the training data as a whole since these were all examples where earlier versions of our model disagreed with the pronunciations in the training data. With heavy dropout and label smoothing as described, early stopping was not required: In particular we did not observe the accuracy on the golden set dropping towards the end of training. In contrast, without such techniques the model would usually start to overfit at about 250K steps, whereas with them the models train to a million steps without overfitting, and still getting higher accuracies.

---

[4]This is not to be confused with shuffling of neighborhoods introduced below in Section 5.2.

## 5 Experiments and evaluation

The various configurations of the model, with and without neighbors, were trained on 2,397,154 neighborhoods, for 1 million steps. Before reporting overall performance results, we illustrate the operation of the with-neighbors transformer model with an example that illustrates the model detecting cases where the data is incorrect. The feature, メゾン日本橋, *Mezon Nipponbashi*, is an apartment building in the Nipponbashi district of Osaka. The problem is that 日本橋 is also a part of Tokyo, pronounced ***Nihon****bashi*, and being more famous, is arguably the "default" pronunciation. The pronunciation of this feature was presumably originally populated by a method that did not take geographical context into account. In Figure 4 we show the feature, the pronunciation as found in the database, the hypothesized (correct) pronunciation, and the neighbors that the model attended to when hypothesizing the feature's pronunciation. The example introduced in Figure 2 is also correctly predicted by the model as *seravi tanmachi*.

In the remainder of this section we present two types of evaluation. First we introduce a non-neural baseline (Section 5.1). In Section 5.2, we present error rates on held-out data for several versions of the model, the non-neural baseline, and a separate RNN model that has been used for more general text-normalization applications. We show that the with-neighbors transformer model has by-far the best performance. In Section 5.3 we delve a bit deeper into the effect of Lat-Long features, as well as details of the performance on the golden set. However, it is also important to show that the model is indeed learning to attend to relevant features in the neighborhood. We present evidence of this in Section 5.4. In Section 5.5 we discuss the important question: How often does a prediction error make the target name incomprehensible?

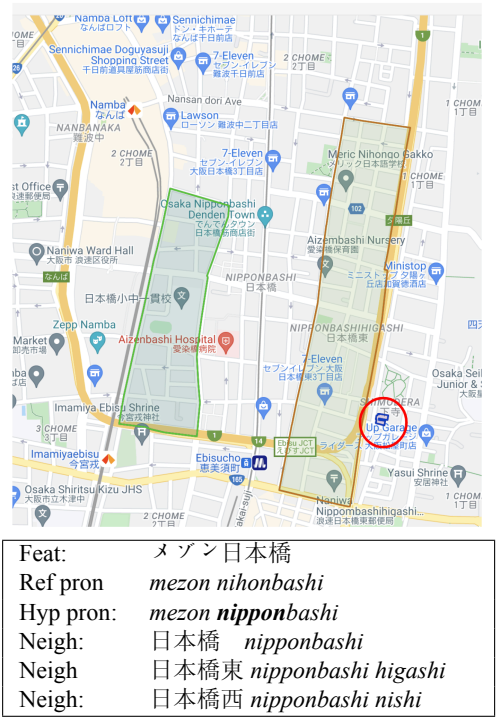

| Feat: | メゾン日本橋 |
| --- | --- |
| Ref pron | *mezon nihonbashi* |
| Hyp pron: | *mezon **nippon**bashi* |
| Neigh: | 日本橋 *nipponbashi* |
| Neigh | 日本橋東 *nipponbashi higashi* |
| Neigh: | 日本橋西 *nipponbashi nishi* |

Figure 4: An error in the original data: メゾン日本橋 *Mezon Nipponbashi* apartments in Osaka, circled in red on the map. Highlighted in green shaded areas are neighboring features, 日本橋東 *Nipponbashi Higashi* and 日本橋西 *Nipponbashi Nishi*.

Moving beyond Section 5, as noted above the maps data has errors, meaning that a small percentage of the cases where the hypothesis of the model differs from what is in the database, the database is in fact incorrect. The main practical application of the model is finding and correcting these sorts of errors. Determining which discrepancies are errors and which are not requires human evaluation, and we report results on this in Section 6. Finally, since manual evaluation is expensive, we would like to be able to decide automatically when we can be confident that a discrepancy should be judged in the model's favor: this is the topic of Section 7. In Section 8, we demonstrate an application of the model to a totally different problem.

## 5.1 Baseline system

As a baseline for comparison we used a proprietary state of the art Japanese text-normalization system to produce pronunciations. The system constructs a lattice using a dictionary and rules, and uses linear models to score paths and Viterbi search to select the best path through the lattice.

This system converts an input feature name to its reading and does not make use of neighbor information. To simulate the use of neighbor information, we first aligned neighbor names with their readings using a kanji-to-hiragana aligner that is part of the text-normalization system introduced above. For example the neighbor name 鹿飼道下 would be aligned to its hiragana reading しかがいみちした as 鹿/しか/*shika*, 飼/がい/*gai*, 道/みち/*michi*, 下/した/*shita*. We then collect statistics on all kanji substrings and their hiragana readings, and keep the most common reading of each substring. Finally, we find the longest span(s) in the target name that match against the substrings collected from the neighbors, and replace the corresponding portion of the name's reading as computed by the text-normalizer, with the reading found from the neighbors. Thus if the text-normalizer produces for 鹿飼道上 the incorrect reading ししかいみちうえ (*shishi kai michi ue*), the method might correct that to しかがいみちうえ (*shika gai michi ue*).

## 5.2 Quantitative evaluation

We evaluated the model on a held-out test set consisting of about 138K neighborhoods, comparing four models: the non-neural baseline (Section 5.1), the with-neighbors and without-neighbors transformer models, and another sequence-to-sequence model, the RoadRuNNer RNN-based neural text normalization system (Zhang et al., 2019) trained on the same data; for RoadRuNNer, the checkpoint with the best string error rate on training was used in evaluation. Note that the RoadRuNNer system has no access to the neighbor information, and thus serves as a baseline sanity check for a sequence-to-sequence model for pronouncing the feature names in the absence of any information about other names in the geographical neighborhood. We also analyze the effects of including Lat-Long embeddings.

We prepared the train-test split in two different ways; in the first, which we refer to as *shuffled* we sample features uniformly across Japan when constructing the two sets. In the second, which we refer to as *unshuffled*, the held out set is actually from non-overlapping areas of Japan such that features in the test set are from areas that the model will not have seen during training. *Clearly the Lat-Long embeddings cannot be used in the latter case since the embeddings for the test area would not be trained.* Here, the point was to verify that the model is still able to generalize by making use of neighbors, in neighborhoods from parts of the country the model will not have seen before. This provides further evidence, in addition to what we discuss in Section 5.4, that the model is learning to use the neighbor information. In practice we use the shuffled set for training and generating corrections in the data (Section 6).

Again, when we speak of error rates on this dataset, we know, as discussed above, that there are incorrect transcriptions, and that therefore there are some cases where the model actually predicts the correct transcription, but is penalized because the ground "truth" contains an error. Nonetheless, while these are frequent enough to be worth using our method to correct them (Section 6), they are still in the minority of cases, and the majority of the time, what is in the data set is correct, which in turn means that one can usefully compare different methods.

Error rates are given in Table 2. For the shuffled data, the error rate of the without-neighbors baseline system (Section 5.1) was 19.9%, which is quite high but reflects the difficulty of the task of reading names of geographical features in Japanese for which the system was not particularly tuned. Using neighbors (see, again, Section 5.1) we can reduce this to 17.9%, a 2 point absolute reduction. While this reinforces the point that neighbors are useful for predicting the pronunciation of a target name, the overall error rates are high. RoadRuNNer outperforms the baseline, with 12.9% error on the shuffled data. The without-neighbors version of the transformer model (10.2%) outperforms RoadRuNNer by 2.7 points absolute, with the with-neighbors transformer reducing the error rate by a further 1.6 points.

Is this reduction significant? Given the Central Limit Theorem for Bernouilli trials (Grinstead and Snell, 1997, p. 330), the 95% confidence interval is given as $\pm\sqrt{p(1-p)/N}$, where $N$ is the number of trials. With $N = 132,753$ and $p = 0.102$ for the without-neighbors transformer model and $p = 0.0862$ for the with-neighbors transformer model, the confidence intervals are $[0.1012, 0.1028]$ and $[0.0854, 0.0870]$ respectively. These do not overlap, suggesting that the differences are significant.

We further compared the two models using paired bootstrap resampling (Koehn, 2004), where for each of the 10,000 trials we randomly with replacement drew $N/2$ elements from the original test set and computed the error rates. This method also indicates the superiority of the with-neighbors model for the nominal significance level $\alpha = 0.05$ with $p < \alpha$ and non-overlapping 95% confidence intervals $[0.104, 0.100]$ for without-neighbors and $[0.088, 0.084]$ for with-neighbors models. Finally, we also confirm the statistical significance by performing the paired permutation test (Good, 2000) using a $t$-statistic, which for 5,000 permutations yields $p = 0.0003$ for $\alpha = 0.05$, where $p < \alpha$.

As expected all the models perform worse on the unshuffled data, since in that case the test data is more dissimilar to the training data, since it is drawn from different regions of the country. Still, the with-neighbors transformer model still gives a significant drop in error rate, reinforcing the point that the model uses neighbors when available.

### 5.3 Lat-Long and Golden Set

Figure 5 shows the effect of adding Lat-Long embeddings for different numbers of neighbors. We see that for the zero-neighbor (= without-neighbors) model the Lat-Long embeddings give a significant boost to the accuracy as one might expect, but as we add more neighbors the benefit appears to diminish, and after about 10 neighbors it seems to hurt performance. This is likely due to overfitting as the extra information makes it easier to memorize a location.

Surprisingly, for the golden set (Table 2, column 5), despite neighbors lowering the error, and the addition of the Lat-Long embeddings lowering it further, the lowest error is achieved by adding Lat-Long embeddings only. We believe that this is due to the different distribution of the examples in the dataset and again the effect of more information allowing overfitting. In practice we keep both the Lat-Long embeddings and use neighbor information for decoding potential correction since the results seem qualitatively better.

### 5.4 The model attends to neighbors

For further confirmation that the model attends to the neighbors, we created artificial data using features containing seven name spellings that have (at least) two pronunciations. To create these seven test sets, we started with *real neighborhoods*, and manipulated them as follows. We focused here on the two most common pronunciations and designated the more common pronunciation as '$P_1$' and the other as '$P_2$'. For example consider toponyms spelled 神戸, primary pronunciation ($P_1$) *koube*, secondary pronunciation ($P_2$) *koudo*. Features containing that kanji spelling include 神戸電池, *koube denchi* 'Kobe Battery', and 神戸大橋 *koudo ōhashi* 'Koudo Bridge'. We decoded our set under three conditions: (1) leaving the pronunciations of the neighbors alone; (2) changing all relevant portions of a neighbor's name, e.g. 神戸 in 神戸電池, to have a $P_1$ (*koube*) no matter what the original pronunciation was; (3) similarly changing all relevant portions to $P_2$, (*koudo*) no matter what the original pronunciation was. The with-neighbors transformer model was then used to decode the target feature, and we measured the proportion of times $P_1$ was decoded under the various (possibly artificial) conditions. The results of this experiment are shown in Table 3. As can be seen in the table, the proportion of $P_1$ is always affected by artificially manipulating the neighbors, though more dramatically so in some cases than others. The signal for the pronunciation *yamato* for 大和 is evidently very strong compared to *taiwa* so that it is very hard to override it with evidence from the neighbors. On the other hand, 日本橋 *nihon/nipponbashi* is easily influenced by the pronunciations of the neighbors. In all cases the neighbors influence the results in the expected direction. This small experiment thus provides further evidence that the model is paying attention to the pronunciations of the neighbors in computing its decision on the pronunciation of a target feature.

| System | ± Neigh | Shuffled | Unshuffled | Golden | Params | Steps |
|---|---|---|---|---|---|---|
| Baseline | − | 0.199 | 0.198 | 0.502 | n/a | n/a |
| Baseline | + | 0.179 | 0.179 | 0.396 | n/a | n/a |
| RoadRuNNer | − | 0.129 | 0.131 | 0.442 | 7.6M | 1M |
| Transformer | − | 0.102 | 0.103 | 0.381 | 6.56M | 1M |
| Transformer | +30 | **0.0862** | **0.088** | 0.367 | 9.74M | 1M |
| Transformer + Lat-Long | − | 0.0892 | n/a | **0.332** | 6.58M | 1M |
| Transformer + Lat-Long | +30 | 0.0867 | n/a | 0.341 | 9.76M | 1M |

Table 2: Error rates for the non-neural baselines, RoadRuNNer (**without neighbors**), the **without-neighbors** transformer model, and the **with-neighbors** transformer model on the test data sets. For the with-neighbors transformer, 30 neighbors were used, hence +30 in the table.

| | | | | Proportion $P_1$ in decoding hypotheses | | |
|---|---|---|---|---|---|---|
| **Name** | **# exx.** | **$P_1$** | **$P_2$** | Original | Neigh. → $P_1$ | Neigh. → $P_2$ |
| 日本橋 | 1,110 | *nihonbashi* | *nipponbashi* | 0.86 | 1.0 | 0.04 |
| 三郷 | 790 | *misato* | *sangou* | 0.79 | 0.83 | 0.06 |
| 佐伯 | 420 | *saeki* | *saiki* | 0.66 | 0.86 | 0.58 |
| 小平 | 780 | *kodaira* | *obira* | 0.91 | 0.93 | 0.62 |
| 神戸 | 4,639 | *koube* | *koudo* | 0.96 | 0.97 | 0.83 |
| 渋谷 | 1,360 | *shibuya* | *shibutani* | 0.98 | 0.99 | 0.85 |
| 大和 | 4,670 | *yamato* | *taiwa* | 0.86 | 0.86 | 0.85 |

Table 3: Synthetic examples demonstrating that the system pays attention to the neighbors. Columns: relevant kanji spelling of the target feature; number of target features; primary pronunciation; secondary pronunciation; proportion of decodings of the primary pronunciation with unchanged data; proportion of decodings of the primary pronunciation when the data are changed as in (2) in the text; proportion of decodings of the primary pronunciation when the data are changed as in (3).

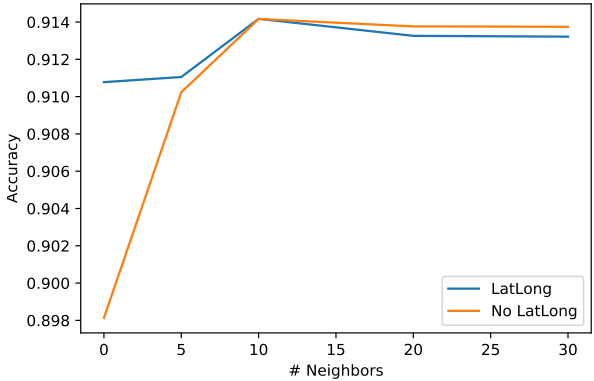

Figure 5: Model accuracies for different number of max neighbors, with and without latitudinal and longitudinal embeddings (shuffled test set).

Further evidence can be seen in visualizations of the transformer attention to neighbors' pronunciations. Figure 6 shows average attention weights over all layers and attention heads. The neighbors from Figure 2 correspond to neighbors 5 and 6 here. When decoding the last four characters たんまち in *tanmachi*, the model is attending to the neighbors that contain this sequence.

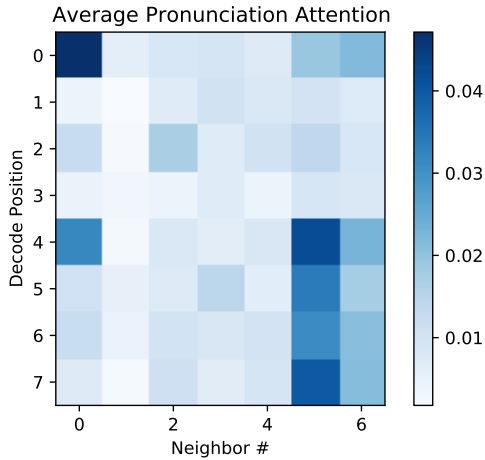

Figure 6: Visualization of transformer attention for the example in Figure 2, with neighbor positions 5 and 6 corresponding to the two neighbors highlighted in that figure. Note the higher attention (darker blue) in the lower right corresponding to たんまち *tanmachi* in the neighbors.

### 5.5 Detailed Error Analysis: How bad are the errors when the model gets it wrong?

As a reviewer for an earlier version of this paper pointed out, reading 日本橋 as *Nihonbashi* as part of a feature name in Osaka (correct pronunciation: *Nipponbashi*) is wrong, but the hearer would likely still be able to understand the intended feature. It should be no worse than reading *Houston (Street)* in New York as /ˈhjuːstən/. A reasonable question is what proportion of the errors that the model makes are similarly 'recoverable' in the sense that the hearer will be able to understand the intended referent. To that end we took a random sample of 60 errors made by the best performing model (with-neighbors transformer model trained on shuffled data, Table 2, row 6) and compared them to the reference transcription from the maps database. The third author, a native speaker of Japanese, evaluated how many of these seemed recoverable in the sense above. Of the 60, 48 were deemed to be recoverable, whereas the other 12 either seemed not to be recoverable or were unclear. An example of a recoverable error is 八幡田, where the reference transcription is *Hachimanda* whereas the model predicts *Hachimanden*. This hinges on the pronunciation of the final kanji using the native (*ta/da*) pronunciation versus the Sino-Japanese (*ten/den*) pronunciation; both pronunciations are in principle possible. Another recoverable example is 比良町, where the reference is *Hirachō*, but the model predicts *Hiramachi*. Again this hinges on the native (*machi*) versus Sino-Japanese (*chō*) pronunciation of the final character. This latter case is particularly hard even for native speakers to get right, since the pronunciation of 町 'town' as *chō* or *machi* is not predictable and must be memorized for each place name.

An example of an unrecoverable error is ロージェ麻生 *Rōje Asao*,[5] which the model predicts as *Rōje Asabu*. In this case, the difference hinges on two native ways to read 麻生, but here the predicted *Asabu* is potentially confusing. While the feature in question is an apartment building in Sapporo, a hearer familiar with Tokyo is likely to confuse it with a well-known area of Tokyo, 麻布 *Azabu*. A more dramatic example is 坤町, a part of Kyoto, where the correct pronunciation is *Hitsujisaruchō*, whereas the model predicted *Konchō*. Once again this hinges on a native (*hitsujisaru*) versus Sino-Japanese (*kon*) pronunciation, in this case for the first character.

So in 80% of the cases, even though the model picks an inappropriate pronunciation, the result is still recoverable. For the remaining 20%, the model did not produce random unrelated pronunciations, but rather theoretically possible pronunciations—indeed errors that a person not familiar with the area might make—but where the pronunciation was deemed too far off to be recoverable. However, we want to stress that in general whether a possible but incorrect pronunciation of a Japanese place name is recoverable or not is an issue that can only be properly answered by a more rigorous study of users in real life situations.

## 6 Finding mistakes in maps data

An important application of the model is to find potential errors in the database, and flag them for possible human correction. To that end, using the with-neighbors transformer model trained with all features, we ran decoding on the *entire* data set, including the training and held-out portions, and identified cases where the model hypothesized a different pronunciation from what was in the reference transcription. In order to focus on the cases of interest, we further filtered these by considering only neighborhoods where some neighbors have spellings that share substrings with the target feature's spelling, and pronunciations that share substrings with the hypothesized pronunciation. This yielded a set of 18,898 neighborhoods that had some discrepancy per the model. Especially for the training portion, it is likely that the model learned whatever pronunciation was in the database, even if it was wrong, so we are likely missing a lot of neighborhoods that have errors: we do not, therefore, know the *recall* of the system. In what follows, we consider the precision, based on a manual analysis by human raters.

---

[5]Apparently for *L'Osier Asao*.

Preliminary analysis of the output revealed that many of the discrepancies involved *establishments*, which include buildings and other man-made features including things like bus stops. These often contain a location name as part of the name. For example, a Family Mart convenience store might be named ファミリーマート小橋駅前店 *Family Mart Kobashi Station Square Store*, with the issue being whether 小橋 is correctly pronounced in the establishment name.

Three raters[6] manually checked pronunciations for 1,056 features, including 555 establishment features. Raters were given links to the feature on Google Maps, and were asked to verify which pronunciation was correct, or give an alternative if neither was correct. Evaluators had to provide 'proof' of their answers, of which the following were considered acceptable: (a) official website of the location, or the Japan Post website; (b) a screenshot from Street View showing the pronunciation (e.g. from a road sign); (c) a Wikipedia page with sufficient appropriate references. Raters were asked not to use other sources.

Overall, the raters found that model correctly detected that there was a potential problem with the reference data 63% of the time. The remaining 37% of the features were actually correct, despite the model having hypothesized a different pronunciation. The 63% of cases with problems broke down as follows: in 36% (absolute) of the cases, the hypothesized replacement pronunciation was correct, in 11% both were wrong (meaning that the model detected a problem, but found the wrong solution) and in 15% of the cases, the rater was unable to verify the answer (which suggests that the feature may need to be checked further). In some categories such as 'compound building', the hypothesized pronunciation was correct (and the reference pronunciation wrong), 80.7% of the time.

---

[6]All raters employed in this study are paid linguistic consultants hired through a third party vendor.

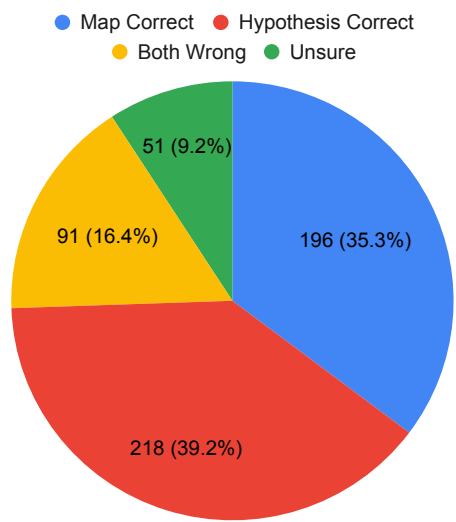

Figure 7: Results of a manual evaluation of 555 establishment features. See the text for an explanation.

Figure 7 shows the results of a manual analysis of the 555 establishments by an independent rater. The establishments represent a range of "impressions", with some appearing frequently in searches, others less so. The rater found that for the majority of cases (55.6%), the data in maps was incorrect: 39.2% where the hypothesized alternative is the correct one; and 16.4% where both what is in the data and the hypothesis are wrong, but where the system has detected a problem with the data. The rater was unsure about a further 9.2%, comprising a further set that should be checked by an expert. (A small percentage of the establishments have closed since the database was created.) Thus only about a third of the establishments selected were correct in the database.

Table 4 gives a breakdown of the two hand-checked samples, considering only cases where either the data already in maps was deemed correct, or the hypothesized replacement was deemed correct. In general the hypothesized corrections had higher accuracies, and the maps data lower accuracies in the establishment set than in the mixed set. Also, the model seems to be making better predictions for the training portion than the held-out portion. Indeed, for the establishments, the hypothesis is more often right for the training portion of the data than what is in the original training data. While the model probably memorizes aspects of the training data, it can still notice discrepancies even in neighborhoods it has seen before.

|  | Mixed sample | | | | |
| --- | --- | --- | --- | --- | --- |
|  | $N$ | +Maps | $\frac{\text{+Maps}}{N}$ | +Hyp | $\frac{\text{+Hyp}}{N}$ |
| Trn | 401 | 149 | 0.37 | 137 | 0.34 |
| Tst | 100 | 49 | 0.49 | 24 | 0.24 |
|  | Establishments only | | | | |
|  | $N$ | +Maps | $\frac{\text{+Maps}}{N}$ | +Hyp | $\frac{\text{+Hyp}}{N}$ |
| Trn | 473 | 161 | 0.34 | 189 | 0.40 |
| Tst | 82 | 35 | 0.43 | 29 | 0.35 |

Table 4: Comparison of two sets of hand-checked features, showing the cases where either the maps data (*+Maps*) *or* the hypothesis were correct (*+Hyp*), broken down into whether the feature in question was in the training, versus the held-out data. $N$ is the total size of each set.

One point that will be clear from the above is that just because there is a discrepancy between the pronunciations of a target feature and the neighboring features does not mean that the target is wrong. Indeed, there are systematic types of features that frequently involve such discrepancies. One such class being train stations, which are notoriously difficult in that they are frequently pronounced differently from the name of the town in which they are located (Imao, 2020). Thus the station that serves 小涌谷 *Kowakudani* is 小涌谷駅 *Kowakidani eki*. Station names were often established during the Meiji Period, and reflect older pronunciations for nearby toponyms.

## 7 Automatic data correction

Unfortunately the model is not yet accurate enough to use it to automatically fix discrepancies for all features. Among the 1,056 manually analyzed features, the original data was correct 37% of the time, and the model 35%, meaning that simply substituting the model's hypothesis would result in a small net loss in accuracy. However, we also saw that the model was more accurate than the reference data for some classes of features, meaning on average the accuracy should increase if we replaced the data in those cases. We have also investigated filtering the data based on metrics extracted from the model itself. For example, we considered decoding entropy as a measure of confidence, the log likelihoods of the beam search outputs, and the relative amount of attention that the attention layers were giving the neighbor summary. Thus far, the most informative measure is difference between the top two beam search decoding log likelihoods. Our interpretation of this is that if there is a large difference in confidence between the two beams then there is little ambiguity in how the model thinks they should be pronounced and thus we can be more confident in the top candidate being correct. Figure 8 shows the ROC curve for using this metric to features from the golden set. It shows an area under the curve of almost 0.7, a clear positive signal, and the Precision-Recall curve shows that we are able to achieve an accuracy of about 80% for about 50% of the data, which still represents a large number of high confidence corrections.

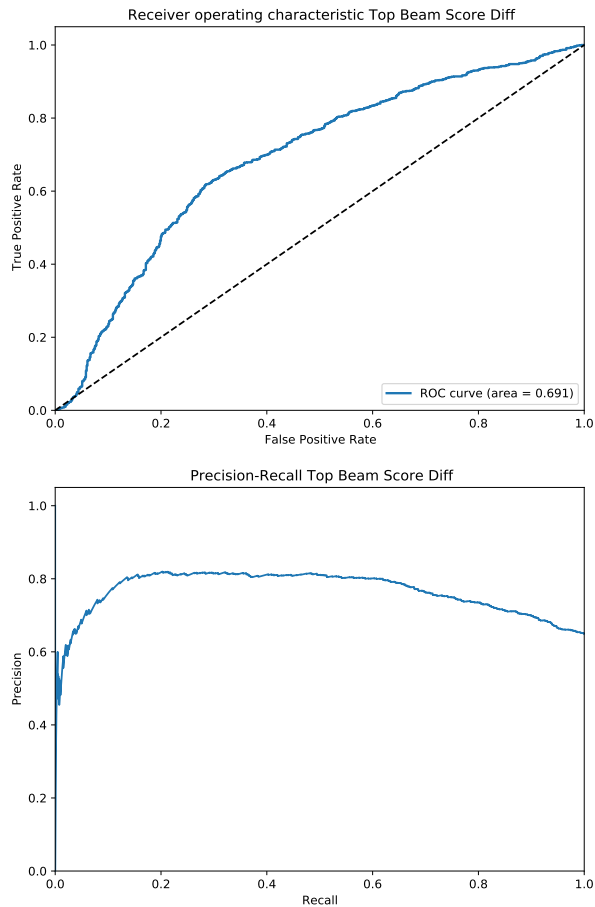

Figure 8: ROC curve (top), and Precision/Recall curve (bottom), for thresholding results on the difference between beam search scores.

## 8 Cognate reflex prediction task

List et al. (2022) present the ACL SIGTYP shared task on the problem of cognate reflex prediction. Cognate reflex prediction is best understood by example. English, Dutch and German are closely related West Germanic languages which share many cognate words. For example English *dream* corresponds to *droom* in Dutch and *Traum* in German. If one now considers English *tree*, the words that correspond in meaning to this in Dutch and German are *boom* and *Baum*, respectively. These are apparently from the same *etymon*, but the English word is not. What should an English cognate look like? On analogy with *dream*, one would predict the form to be *beam*. Indeed, while *beam*'s meaning has shifted, it is in fact related to *boom* and *Baum*. In the SIGTYP task, participants were presented with data from several language families, where the task was to reconstruct what the cognate forms for particular etyma would be, given examples in a subset of the sister languages.

| Cognate reflex | | Geo. name task |
|---|---|---|
| target form | /dare/ | ← main feature$_1$ pron. |
| target lang. | L$_1$ | ← main feature$_1$ name |
| related form | /θaru/ | ← neigh. feature$_2$ pron. |
| related lang. | L$_2$ | ← neigh. feature$_2$ name. |
| related form | /dare/ | ← neigh. feature$_3$ pron. |
| related lang. | L$_3$ | ← neigh. feature$_3$ name. |
| related form | /daɨ/ | ← neigh. feature$_4$ pron. |
| related lang. | L$_4$ | ← neigh. feature$_4$ name. |
| ⋮ | ⋮ | ⋮ |

Table 5: Parallels between the cognate reflex prediction and the geographical name reading prediction tasks. "L$_1$" and so forth in the first column represent the names of the languages in the set.

| System | Rank | NED | B-Cubes | BLEU | Aggregated |
|---|---|---|---|---|---|
| Inpainting | 1 | 1 | 1.2 | 1 | 1.1 ± 0.3 |
| **Neighbors 30K** | 2 | 2.6 | 3 | 2.6 | 2.7 ± 0.4 |
| **Neighbors 35K** | 3 | 2.4 | 4 | 2.4 | 2.9 ± 0.9 |
| SVM Baseline | 4 | 5.2 | 4 | 5 | 4.7 ± 1.9 |
| Neighbors 100K | 5 | 4.6 | 6.6 | 4.6 | 5.3 ± 1.3 |
| System 2 | 6 | 6 | 7 | 6.2 | 6.4 ± 1.1 |
| System 3 | 7 | 7.6 | 4 | 7.6 | 6.4 ± 2.5 |
| CORPAR Baseline | 8 | 6.8 | 6.2 | 6.8 | 6.6 ± 0.8 |
| System 4 | 9 | 8.8 | 9 | 8.8 | 8.9 ± 0.4 |

Table 6: Average ranks of systems in the SIGTYP 2022 Shared Task along with aggregated ranks.

Kirov et al. (2022) report the results of applying two models to this task, one being a model based on *image inpainting* (Liu et al., 2018), and the second being a variant of the neighbors model presented in this paper. The cognate reflex prediction problem is similar in spirit to the geographical feature reading task, where we replace "neighbor reading" with the form of a cognate in a related language, and "target reading" with the form to be predicted. As for the "spellings", we replace these with a string representing the name of the language associated with each of the neighboring cognates and with the target. Table 5 summarizes the parallels between the two tasks. The model used by Kirov et al. (2022) differed slightly from the version reported above in that the language identifiers and cognate forms are interleaved and then concatenated together and attended to directly by the decoder without any averaging, and *source token* ids are added to each cognate in the set. This allows the model to better attend to the individual cognate and to copy (portions of) the cognate as needed. Also, since the data sets for the cognate reconstruction task are small, a smaller transformer configuration was used. Even so, the provided data sets were too small, so Kirov et al. (2022) augmented the data in two ways. First, the data were augmented by copying neighborhoods while randomly removing neighbors, thus making new neighborhoods for the same cognate set. Second, synthetic cognate sets were generated for each of the "neighbor" languages and the target using simple $n$-gram models trained on the provided data.

The two systems developed by Kirov et al. (2022) achieved the top ranking in the shared task. In general the better performing of the two was the inpainting model, but on some language families, such as Semitic, the neighbors model outperformed the inpainting model. Table 6, adapted from (List et al., 2022, Table 4), shows the results for two baselines, the inpainting model, three versions of the neighbors model—30K, 35K and 100K training steps—and three other competing systems. The rank in the final column is aggregated over the *normalized edit distance* (NED), *B-cubed F-scores* and *BLEU*. The inpainting model and the neighbors model were the only two systems that overall outperformed the SVM baseline. The fact that the neighbors 30K neighbors model worked better than higher numbers of training steps can likely be attributed to overtraining. These results suggest that the model we have presented in this paper has potential applications outside the main task we have reported here.

## 9 Discussion

In this paper we have presented a novel architecture for the difficult task of pronouncing Japanese geographical features that learns to use pronunciations of features in a local neighborhood as hints. We have shown via various means that the model pays attention to the neighboring features, and that therefore the model has learned what we intended it to learn: that in order to pronounce a name, it is often useful to consider how neighbors are pronounced. We also conducted manual evaluations showing that for some classes of features, model hypotheses differing from pronunciations in the database could be as high as 80% correct. Our results are currently being used to correct errors in Google Maps. In future work we also plan to extend the coverage of the model beyond Japan. While Japanese place names are particularly difficult, we noted in the introduction that there are similar problems in other regions. One problem that comes up in the United States, for example, is nonce abbreviations for certain features. For example if one looks in Google Maps in Shreveport, LA, one will run across the weirdly abbreviated *Sprt Bkdl Hwy Srv Dr*. Out of context this is virtually uninterpretable, but if one looks at nearby features one will find the *Shreveport Barksdale Hwy*. From this and other information one can deduce that the mysteriously named feature must be the *Shreveport Barksdale Highway Service Drive*.

Besides geographical names, there are other problems to which a similar approach can be applied. The neighbor model can be thought of as an auxiliary *memory*, to be consulted or not depending on the decision being made. We discussed one possible application of this conceptualization to the task of cognate reflex prediction in Section 8.

A further extension of the idea is *joint* correction of features in a neighborhood. If most of the pronunciations of 日本橋 in a neighborhood are *nipponbashi*, then one could consider correcting all cases where the pronunciation is listed as *nihonbashi* in the neighborhood, not just the main feature. Note that this is somewhat similar in spirit to work on *collective classification* (Sen et al., 2008).

Finally, it is also worth noting that while our work has been with a proprietary maps database, there are open-source maps datasets such as OpenStreetMap (Haklay and Weber, 2008), which likely have at least as many problematic issues as the database we used. The techniques we describe in this paper could be applied to improving such data.

## 10 Acknowledgments

We thank three anonymous reviewers of previous versions of this paper for detailed feedback. We also thank Jesse Rosenstock for help with the code that extracts neighborhoods.